\documentclass{article}
\usepackage[preprint]{corl_2026} %

\usepackage{amsfonts}
\usepackage{amsmath}
\usepackage{multirow}
\usepackage{booktabs}
\usepackage{graphicx}
\usepackage{array}
\usepackage{tabularx}
\usepackage{subcaption}
\usepackage{xcolor}
\usepackage{makecell}
\usepackage{float}
\usepackage{listings}
\usepackage{amssymb}
\usepackage{algorithm}
\usepackage{algpseudocode}

\newcolumntype{Y}{>{\raggedright\arraybackslash}X}

\newcommand{\ours}{\textsc{Feijoa}}
\newcommand{\fido}{\textsc{Fido}}
\newcommand{\llm}{\textsc{LLM}}

\newcommand{\tlp}[1]{}

\begin{document}
\raggedbottom  %
\setlength{\textfloatsep}{12pt plus 2pt minus 4pt}  %
\makeatletter
\renewcommand{\section}{\@startsection{section}{1}{\z@}{-1.5ex \@plus -0.5ex \@minus -0.2ex}{1.0ex \@plus 0.3ex \@minus 0.2ex}{\large\bf\raggedright}}
\renewcommand{\subsection}{\@startsection{subsection}{2}{\z@}{-1.3ex \@plus -0.5ex \@minus -0.2ex}{0.5ex \@plus 0.2ex}{\normalsize\bf\raggedright}}
\renewcommand{\subsubsection}{\@startsection{subsubsection}{3}{\z@}{-1.0ex \@plus -0.5ex \@minus -0.2ex}{0.3ex \@plus 0.2ex}{\normalsize\bf\raggedright}}
\makeatother

\title{\LARGE \bf \ours{}: Task-Driven Grounding and Execution for Mobile Manipulation}

\author{
  \textbf{Aditya Agarwal}$^{*}$ \quad \textbf{William Shen}$^{*}$ \quad
  \textbf{Sahit Chintalapudi}$^{\dagger}$ \quad \textbf{Lucy Cai}$^{\dagger}$ \quad
  \textbf{Nishanth Kumar} \\
  \textbf{Aidan Curtis} \quad \textbf{Travis Manderson} \quad
  \textbf{Tom\'as Lozano-P\'erez} \quad \textbf{Leslie Pack Kaelbling} \\
  MIT CSAIL \\
  \url{https://williamshen-nz.github.io/feijoa/} \\
  \vspace{2pt}
  {\footnotesize $^{*}$Equal contribution. \quad $^{\dagger}$Equal contribution.}
}

\maketitle
\begin{figure}[h]
\centering
\includegraphics[width=\linewidth]{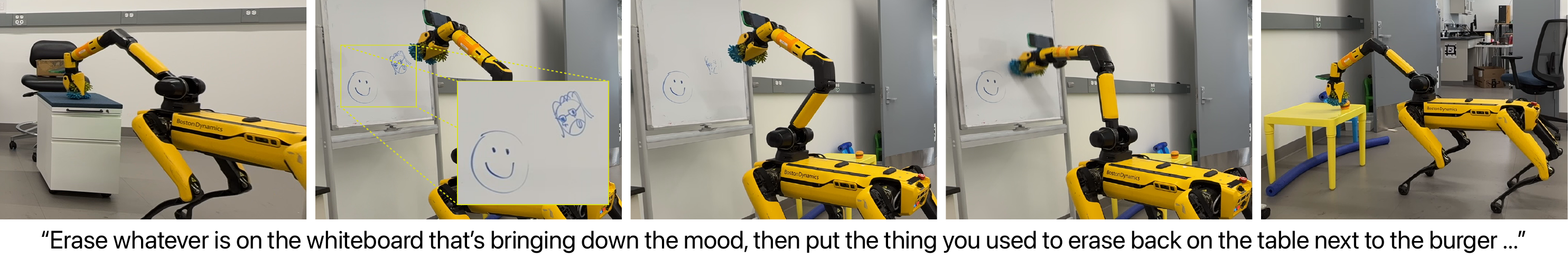}
\caption{\textbf{\ours{} on Spot.} Given ``Erase whatever is on the whiteboard that's bringing down the mood, then put the thing you used to erase back on the table next to the burger'', the agent must work out that the drawing to remove is the sad face, that the eraser is the mitt it is holding, and which of several tables has the toy burger on it. Left to right: it grounds and picks up the mitt, drives to a base pose from which the board is within reach, wipes the face away, checks from the before and after images that the marks are gone, and returns the mitt to the table beside the burger.}
\label{fig:teaser}
\end{figure}

\begin{abstract}
A mobile manipulator asked to ``put the toys away in the toy box'' must decide which objects are toys and which container the user means, then find base poses, grasps, and placements it can execute.
Because a scene representation is built before the instruction is known, it may lack the attributes, affordances, or concepts needed to resolve the task.
We present \ours{}, a language-model agent that grounds the instruction at test time. It queries a grounding tool over a posed RGB-D scan, inspects the returned candidates, and refines its queries until they resolve the instruction.
An embodiment-aware harness checks each skill's preconditions, samples base poses from which the target is reachable, and verifies the skill's effect from observations before and after execution.
On 103 open-ended grounding queries, \ours{} resolves 94.2\% exactly, compared with 68.0\% for the best prebuilt representation.
Over 50 real-world trials on Spot, it completes 35 tasks; 17 of these successes follow recovery from a failed tool call or verification check. Of the 15 failed trials, 13 were caused by the robot's physical skills, and only 2 by grounding or the agent's reasoning. 

\end{abstract}

\section{Introduction}
\label{sec:intro}

Robots operating in homes and offices must execute long-horizon tasks from open-ended natural language instructions.
Consider asking a robot to ``Clean up the room so the plush toys are all on the same table''. 
To complete this task, the robot must identify all the plush toys among everything else in the room, determine which table the user likely wants the toys placed on (one table already has four plush toys on it), navigate to base poses from which it can grasp and stably place the toys onto the table, and monitor its progress and recover from failures in execution.
More generally, this requires open-ended grounding that depends on the task, the state of the environment, and common-sense reasoning. Acting on what is grounded requires finding base poses, grasps, and placements that are collision-free and feasible for the robot's embodiment.

Many existing approaches construct task-agnostic semantic scene representations before the instruction is known. These include open-vocabulary 3D object maps~\cite{morin2025agentic}, 3D scene graphs encoding spatial and semantic relationships between objects~\cite{rana2023sayplan,gu2024conceptgraphs,maggio2024clio,yan2025dovsg}, and 3D feature fields~\cite{jatavallabhula2023conceptfusion,kerr2023lerf,shen2023distilled}.
These representations make large environments efficient to index, since the expensive perception is done once, before any query.
However, they reduce the scene to a collection of embeddings or captions, and are therefore incomplete with respect to any specific instruction: they may lack the fine-grained attributes, affordances, and concepts it requires (e.g., ``most colorful object'', ``plush toy with a bow tie'', ``cleanest table with shelves above it'').
We argue that for open-ended mobile manipulation, grounding must be \emph{task-driven} and computed against raw observations at \emph{test time}: which objects to find, and what relationships to resolve between them, depend on the instruction.

\begin{figure}[t]
\centering
\includegraphics[width=0.95\linewidth]{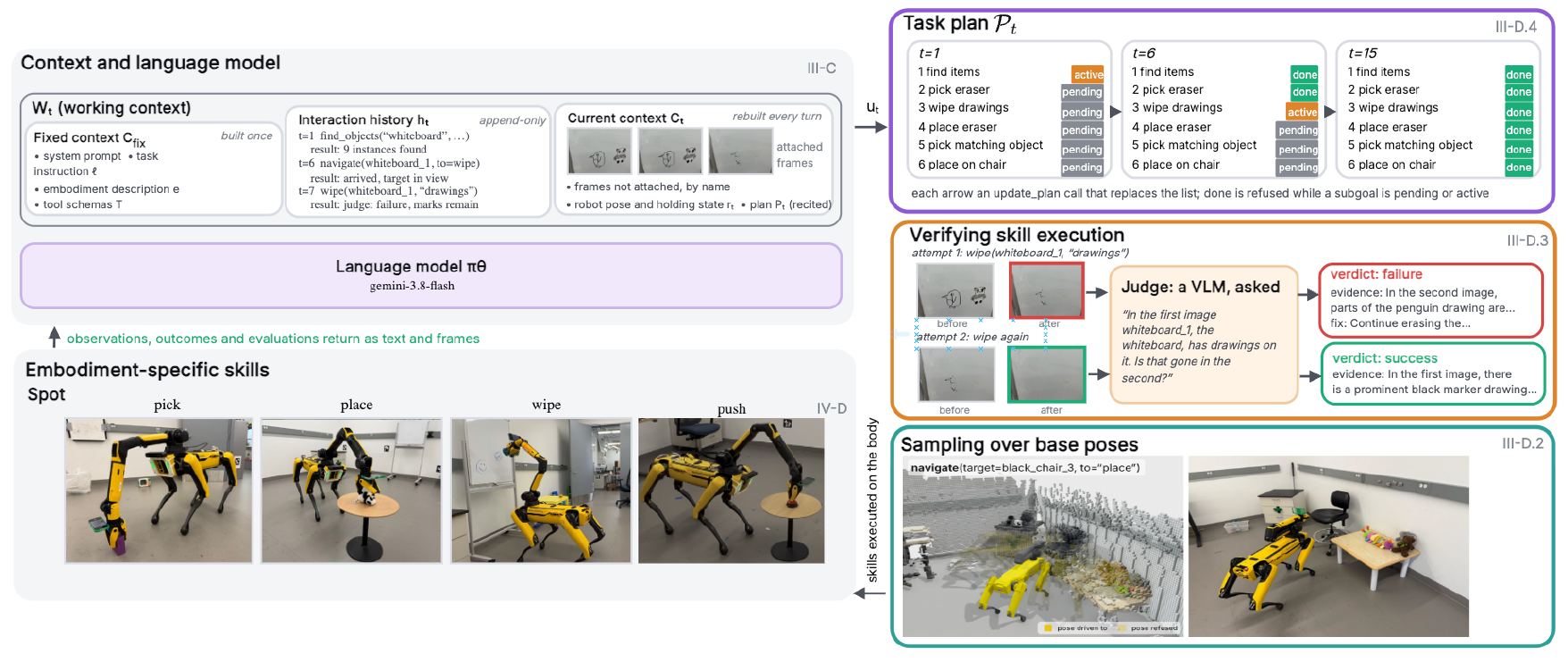}
\caption{
\textbf{\ours{} overview.} At each turn the harness assembles the working context $\mathcal{W}_t$ from a fixed context, the interaction history $h_t$, and the current context and the task plan $\mathcal{P}_t$. \llm{} $\pi_\theta$ selects tool calls $u_t$, which the harness executes. \fido{} (\ours{} grounding subagent) generates object hypotheses over the recorded scan. Each skill requires feasibility-aware sampling, precondition checks, and effects verified. Skills execute on the embodiment through robot-specific implementations, and results, failures, verdicts, and new observations are appended to $h_t$ for the next decision.}
\label{fig:method}
\end{figure}

We introduce \ours{}, a language-model agent for long-horizon mobile manipulation that decides what to ground and how to act (Figure~\ref{fig:method}).
The agent has access to tools that ground objects, update and mark off subgoals in its task plan, estimate the cost of navigating between objects, and execute robot skills such as navigation, picking, placing, pushing, and wiping.
At each turn, the agent chooses among these tools based on the current observations and robot state, its task plan, and the history of its reasoning and tool results.
Our embodiment-aware harness constructs this context for each turn, dispatches every tool call, and records its result for the agent's next decision.
For a robot skill, the harness also checks feasibility before execution and verifies the effects afterwards.

We evaluate \ours{}'s grounding performance, ablate key design decisions in the harness, and deploy it in the real world on a Spot quadruped with an arm and an RB-Y1 bimanual humanoid with a wheeled base.
Over 103 grounding queries that reference objects by category, attribute, spatial relation, membership in a set, or that require reasoning beyond appearance (e.g., negation, affordance, comparison), \ours{} finds every target object with no false positives on 94.2\% of them.
\ours{} significantly outperforms the best baseline, a text-captioned 3D scene graph at 68.0\%, and our grounding tool without the agent at 47.6\%.
We demonstrate that each design decision in our harness matters by ablating effect verification, feasibility-aware sampling, and cost estimation.
Finally, we run 50 real-world trials on Spot over short, medium, and long horizon tasks, in which the agent made 1,045 tool calls in total and executed up to 12 manipulation skills per trial, excluding navigation.
\ours{} succeeds on 35/50 trials, 17 of them after recovering from a failed tool call or verification check.
Of the 15 failed trials, 13 were caused by the robot's physical skills, and only 2 by grounding or the agent's reasoning. 

Overall, we make three contributions: (1) task-driven grounding, in which a language-model agent decides what and how to query our grounding tool and judges its results against the instruction;
(2) an embodiment-aware agent harness that checks a skill's preconditions and samples feasible parameters before execution, and verifies the effects afterwards;
and (3) 50 real-world mobile manipulation trials on Spot with an analysis of how the agent recovers from failed tool calls and where the system fails.

\section{Related Work}

\textbf{Semantic scene representations.}
A central question in robot perception is what information to encode about the environment and at what granularity. Open-vocabulary 3D representations associate spatial locations with foundation-model embeddings, either as implicit or fused dense maps~\cite{shafiullah2023clipfields,jatavallabhula2023conceptfusion,peng2023openscene} or as language-aligned feature fields~\cite{kerr2023lerf,shen2023distilled}. These per-point representations support open-vocabulary retrieval but lack explicit object-level structure.
Object-centric representations instead lift 2D detections into 3D object maps with associated semantics, supporting open-vocabulary retrieval, pick-and-place, and online updates for dynamic scenes~\cite{lu2023ovir3d,liu2024ok-robot,liu2024dynamem}. Scene graphs additionally encode inter-object relationships, built hierarchically or by fusing 2D foundation-model outputs across views~\cite{hydra,gu2024conceptgraphs,koch2024open3dsg,Werby-RSS-24-hierarchical-ovsg,yan2025dovsg}. Clio~\cite{maggio2024clio} selects granularity from a task description via an information-theoretic criterion, and MomaGraph~\cite{ju2025momagraph} encodes functional relationships and part-level interactive nodes for embodied planning.
All of these are encoded before the task is known, so fine-grained attributes, transient states, and task-specific affordances may be absent. Moreover, affordances also depend on the robot's skills, making them difficult to precompute exhaustively. \ours{} sidesteps this trade-off by extracting task-relevant information on demand from raw observations.

\textbf{Agents for embodied task execution.}
Language models have been used as reasoning engines that interleave reasoning with action selection~\cite{yao2023react}, generate programs over vision APIs~\cite{suris2023vipergpt}, and write robot policy code or 3D value maps for manipulation~\cite{liang2023code-as-policies,huang2023voxposer}. Early LLM-based robot planners generate action sequences directly from instructions over flat action spaces~\cite{huang2022languagemodels-zeroshot-planner}, grounded by affordance functions~\cite{ahn2022saycan}, closed-loop feedback~\cite{huang2022inner-monologue}, or few-shot visual grounding~\cite{song2023llmplanner}. SayPlan~\cite{rana2023sayplan} scales to large environments by planning over prebuilt 3D scene graphs, so plan quality is bounded by what the graph encodes. 
More recent systems use language models as closed-loop agents that compose perception and action tools~\cite{feng2025reflective,shi2025maestro}. Agentic Scene Policies (ASP)~\cite{morin2025agentic} is concurrent work equipping an LLM agent with tools over a 3D object map and predefined spatial query functions. \ours{} differs in two respects: it grounds each reference against the raw scan when the instruction arrives rather than querying a map built beforehand, and it gates each skill on its preconditions, on sampling a base pose from which the skill is feasible, and on verifying its effect from fresh observations. We evaluate on long-horizon mobile manipulation on a real robot, where these checks determine whether a plan survives execution.

\section{Problem Formulation And Approach}
\label{sec:problem_and_approach}

\subsection{Problem Definition}
\label{sec:problem_definition}
We consider a mobile manipulator tasked with executing a natural-language instruction $\ell$ in a previously scanned workspace. The agent is given an initial posed RGB-D recording $\mathcal{D}_0 = \{(I_i, D_i, K_i, T_i)\}_{i=1}^{N}$, which serves as its \textbf{Observation Memory}, where $I_i$ and $D_i$ are the RGB image and depth map, $K_i$ denotes the camera intrinsics, and $T_i \in SE(3)$ maps camera coordinates into a common world frame. The robot is localized in this frame. 
Let $e$ denote the robot embodiment and $\mathcal{K}_e$ its library of parameterized skills, whose applicability depends on the robot's kinematics and footprint relative to the scene's geometry. During execution, the agent can acquire fresh observations $o_t$ and access robot measurements $r_t$, including its current pose and holding state. We assume that the environment changes only through the robot's actions; consequently, the initial recording provides prior observations that must be updated as execution proceeds. The instruction does not explicitly identify the corresponding objects in the scene or specify how the robot should physically interact with them and must be inferred from the instruction and available observations.
The objective is to select perception queries and parameterized skills that accomplish $\ell$ while respecting the embodiment's physical constraints, minimizing expected travel cost. We denote successful completion by $\operatorname{Sat}(s_T,\ell)=1$, where $s_T$ is the final physical state and $\operatorname{Sat}$ expresses whether that state satisfies the instruction $\ell$.
We limit execution to $H$ interaction steps and a monetary inference budget $C_{\max}$. $\operatorname{Sat}$ is not available to the agent as an executable goal test.

\subsection{Working Context, Tools, and Decision Loop}
\label{sec:problem_agent}

\textbf{\ours{} agent}: We address this problem through \ours{}, a \textbf{history-conditioned agent} comprising an off-the-shelf language model (\llm{}) and a robotics harness. We use Gemini-3.8-Flash with medium reasoning effort as the \llm{}. The \llm{} performs high-level reasoning and selects tool calls and their arguments from a working context $\mathcal{W}_t$. We build the robotics harness as a closed-loop orchestrator around this \llm{}, that constructs this context, maintains the task plan, exposes the tools that \llm{} can invoke, and coordinates tool execution, and feedback for subsequent decisions.

We provide a tool library $\mathcal{T}=\{\tau_1,\ldots,\tau_M\}$, consisting of grounding, perception, planning, and robot tools where each tool has a description and an argument schema. For grounding, \texttt{find\_objects} returns object hypotheses matching language queries over $\mathcal{D}_0$. The perception tools \texttt{observe} and \texttt{fetch\_frames} acquire fresh robot-camera images and retrieve previously stored images by frame identifier, respectively. For planning, \texttt{estimate\_cost} compares estimated navigation costs for different visit orders proposed by the \llm{}, and \texttt{update\_plan} revises the task plan. The \texttt{ask\_human} tool waits for operator input and returns it as tool feedback; \texttt{done} and \texttt{fail} declare completion or failure. We expose navigation and manipulation through \texttt{navigate}, \texttt{pick}, \texttt{place}, \texttt{push}, and \texttt{wipe}, using selected object hypotheses and skill-specific arguments. We describe grounding and skill execution in Secs.~\ref{sec:problem_grounding} and~\ref{sec:problem_action_selection}.
At each step, the harness assembles $\mathcal{W}_t = \operatorname{BuildContext}(\mathcal{C}^{\mathrm{fix}}, h_t, \mathcal{C}_t^{\mathrm{cur}})$ from three components,
where the fixed context $\mathcal{C}^{\mathrm{fix}}$ comprises a minimal system prompt, embodiment description, task instruction $\ell$, and the available tools $\mathcal{T}$.
The interaction history $h_t$ is an append-only textual record of the \llm{}'s previous messages, tool invocations with their arguments, returned results, and references to recorded observations.
The current context $\mathcal{C}_t^{\mathrm{cur}}$ is rebuilt every step using selected annotated grounding images and current or previous robot observations, and robot measurements $r_t$ including its pose and what is held, and the latest task plan $\mathcal{P}_t$. 
This context serves as the agent's internal memory.
At each decision step, the language model selects tool invocations and their arguments from the set of available tools $\mathcal{T}$ according to $u_t \sim \pi_\theta(\cdot \mid \ell, e, \mathcal{W}_t)$,
where $\pi_\theta$ denotes the language model's conditional distribution over tool invocations, and $u_t$ contains the generated tool calls and their arguments, which are then executed by the harness. The returned results include the intermediate computations, execution outcomes, and supporting observations $o_t$ when available. The harness appends textual results and observation references to $h_t$ and builds the next working context $\mathcal{W}_{t+1}$, which the \llm{} uses to make subsequent decisions.

To track progress and support recovery from failures in long-horizon tasks, the harness maintains a persistent task plan $\mathcal{P}_t$. The \llm{} is prompted to construct an initial task plan at the beginning of the episode from \(\ell\) and $\mathcal{W}_t$. As the task progresses, the \llm{} updates the plan through \texttt{update\_plan()} using new observations and execution feedback to record progress and revise the remaining task. The harness validates these updates before storing them in $\mathcal{P}_t$. We represent the task plan as $\mathcal{P}_t =[(g_j,\chi_j,\sigma_j)]_{j=1}^{m_t}$
where $m_t$ is the number of subgoals, $g_j$ describes a subgoal, $\chi_j$ describes what counts as completing it, and $\sigma_j$ records its language-model-assigned status.
The harness always retains the latest $\mathcal{P}_t$, and includes it in the $\mathcal{C}_t^{\mathrm{cur}}$ at every decision step.

\subsection{Task-Conditioned Grounding}
\label{sec:problem_grounding}

Given a task instruction $\ell$ and a working context $\mathcal{W}_t$ defined in Sec.~\ref{sec:problem_agent}, our grounding subagent \fido{} performs task-conditioned grounding by resolving language queries into a set of \textbf{object hypotheses}.
Given $\ell$, the \llm{} first formulates one or more grounding queries ($\mathcal{Q}_t = \{q_{t,i}\}_{i=1}^{L_t}$, where $L_t$ is the number of queries) describing the objects to search for. 
\fido{} resolves each query $q_{t,i}$ by invoking the
grounding operator $\mathcal{G}$ over the observation memory $\mathcal{D}_0$ (Figure~\ref{fig:grounding}) as $\mathcal{Z}_t = \mathcal{G}(q_t,\mathcal{D}_0) = \{z_j\}_{j=1}^{n_t}$.
$\mathcal{G}$ returns a set of $n_t$ \textbf{object hypotheses} $\mathcal{Z}_t$, where each individual hypothesis $z_j=(\mathrm{id}_j,\hat{\mathbf p}_j,\hat{\mathbf d}_j,\mathcal V_j)$ represents a possible object matching the grounding query, with an identifier $\mathrm{id}_j$ for subsequent tool calls, estimated 3D position $\hat{\mathbf p}_j$ and spatial extent $\hat{\mathbf d}_j$, and supporting visual observations $\mathcal{V}_j$.

For example, given $\ell$ ``put the plush toys into the bin'', \fido{} requests $\mathcal{G}$ for ``plush toys'' ($q_1$) and ``bin'' ($q_2$), and these queries return two toy hypotheses, $\mathcal{Z}^{\mathrm{toy}} = \{z_1,z_2\}$, and one bin hypothesis, $\mathcal{Z}^{\mathrm{bin}} = \{z_3\}$.
\fido{} can compose multiple queries, compare their outputs, and revise grounding queries to resolve attributes, relationships, or ambiguity. 
For example, for the instruction ``put the plush toy closest to the chair into the bin'', the \fido{} can request grounding for the plush toys, chair, and bin; inspect the returned instances and their estimated positions, compare their spatial relationships, and select the plush toy closest to the chair to put into the bin; requesting further information if needed. 
\fido{} can then use these identifiers $\mathrm{id}_j$ to parameterize subsequent navigation and manipulation skills. We provide \fido{}'s implementation details in the Appendix.

\subsection{Action Selection and Closed-Loop Execution}
\label{sec:problem_action_selection}
\subsubsection{Skill Parameterization}
\label{sec:problem_skill_parameterization}
For each robot embodiment $e$, we define a library of parameterized robot skills $\mathcal{K}_e$. Each skill is parameterized via an object hypothesis's $z_j$ parameters $(\mathrm{id}_j, \hat{\mathbf p}_j, \mathcal V_j, k)$, a query label $q$, and a skill-specific language instruction $\eta$ where applicable. 
Each robot skill uses $(\mathrm{id}_j,\hat{\mathbf p}_j)$ from the selected object hypothesis $z_j$. The additional inputs are $k$ for \texttt{navigate}; $(\mathcal V_j,\eta_{\mathrm{pick}})$ for \texttt{pick}; $(q(z_j),\eta_{\mathrm{place}})$ for \texttt{place}; $q(z_j)$ for \texttt{push}; and $(q(z_j),\eta_{\mathrm{wipe}})$ for \texttt{wipe}. The instructions $\eta_{\mathrm{pick}}$ and $\eta_{\mathrm{place}}$ are optional.
We associate each skill with its preconditions $\operatorname{Pre}_{k, e}$ and intended effects $\operatorname{Eff}_{k, e}$. The preconditions specify the visual observation and geometric constraints which includes the skill-specific kinematic requirement $R_{k, e}$ that must be satisfied by the $e$ before a skill can be attempted. The intended effects describe the outcome we seek from execution. We use these geometric constraints  when sampling for continuous parameters (Sec.~\ref{sec:problem_sampling}), and verify skill outcomes using a judge against the updated visual observation (Sec.~\ref{sec:problem_execution}).

\subsubsection{Sampling Over Continuous Parameters}
\label{sec:problem_sampling}
We derive a two-dimensional occupancy grid $\mathcal M_e$ from the observation memory $\mathcal D_0$, using the robot's height. We use $\mathcal M_e$ with the skill's kinematic requirements $R_{k, e}$ to determine where the robot can stand when attempting the skill. 
For a selected skill $k$ and grounded target $z_j$, we sample a set of candidate base poses $\mathcal B$, where each pose $\mathbf b=(x,y,\psi)$ faces the target's estimated 3D centroid $\hat{\mathbf p}_j$.
We retain poses whose footprints lie within free space
and satisfy the skill's kinematic requirements:
\[
    \widehat{\mathcal B}
    =
    \left\{
        \mathbf b \in \mathcal B
        \;\middle|\;
        \begin{array}{l}
            F_e(\mathbf b,\mathcal M_e)=1,\\
            R_{k,e}(\mathbf b,\hat{\mathbf p}_j)=1
        \end{array}
    \right\},
\]
where $F_e$ checks for collision between footprint and $\mathcal M_e$ and $R_{k,e}$ checks whether the embodiment $e$ can perform the task $k$ kinematically. We rank the retained poses using reach margin and estimated travel cost, and plan navigation to the selected pose. If no candidate passes the checks, or navigation planning to the selected pose fails, the tool returns a refusal explaining the failure.

\subsubsection{Verifying Skill Execution Outcome}
\label{sec:problem_execution}

After a skill attempt, we assess its intended effect $\operatorname{Eff}_{k,e}$ using the skill execution status, robot state, and visual observations. For example, gripper feedback provides evidence that an object is being held, while the base pose indicates where the navigation ended. For skills assessed visually, we use a visual judge $J$ as an assessment function. For each verification call, the skill implementation constructs a claim $c_t$ from its intended effect $\operatorname{Eff}_{k,e}$, and the grounded target $z_j$, and any additional parameters $c_t$ as $c_t = \operatorname{Eff}_{k,e}(z_j, \eta)$. For example, a placement claim verifies whether previously held object is at its intended destination. The skill implementation provides visual observations before and after the attempt, denoted as $o_t^{-}$ and $o_t^{+}$ respectively. 
We define the judge's output as $(v_t,\epsilon_t,\rho_t) = J(c_t,o_t^{(1)},o_t^{(2)})$,
where $v_t \in \{\texttt{success},\texttt{failure}, \texttt{uncertain}\}$ is the verdict, $\epsilon_t$ is its textual justification, and $\rho_t$ is an optional suggested correction. We set $\rho_t=\varnothing$ when no correction is provided. We implement $J$ using a separate Gemini-3.6-Flash model with a fresh context containing just the claim and selected observations, and constrain its output to the structured verdict above. We assess each skill against its own intended effect, and return the judge's verdict, and robot state, along with observations as feedback for subsequent agent decisions.

\subsubsection{Feedback and Plan Revision}
\label{sec:problem_feedback_revision}
After tool execution, we record the feedback into \(h_t\), which includes skill execution status, robot status $r_t$, new observations, and the judge's $J$ verdict. From the updated working context $\mathcal W_{t+1}$, \ours{} decides the next tool invocation using the same history-conditioned policy.

Execution failures can include cases where the requested skill did not proceed because a required precondition or feasibility check fails, or the judge $(v_t,\epsilon_t,\rho_t)$ declares a failure with feedback. 
On successful execution, the agent uses the feedback to record task progress and move on to the next subgoal, and revisits the remaining steps in $\mathcal{P}_t$. 
A refusal does not necessarily change the task plan. The agent may resolve the immediate issue while keeping the same subgoal. The plan changes only when the agent explicitly calls the \texttt{update\_plan}.
The agent keeps runs in a closed-loop until the task is achieved, or the step or cost budget is exhausted, or it explicitly declares a failure.

\section{Experiments}
\label{sec:experiments}

We design our experiments to answer the following questions: (i) How does \textbf{grounding} task-relevant objects directly from the RGB-D scan at query time perform against methods that ground over representations built before the query? (ii) Which design decisions in the \textbf{execution harness} matter for reliable execution? (iii) How does the full system perform end-to-end on $50$ real-world mobile manipulation tasks and what are the primary \textbf{failure modes}.

\subsection{Open-Ended Grounding}
\label{sec:exp_grounding}

\begin{table*}[t]
\centering
\caption{\textbf{Open-Ended Grounding.} Results on 103 queries across four scans. Query time is mean $\pm$ std, excluding preprocessing; cost reflects hosted model usage.}
\label{tab:grounding}
\footnotesize
\setlength{\tabcolsep}{3pt}
\renewcommand{\arraystretch}{1.15}
\begin{tabular*}{\textwidth}{@{\extracolsep{\fill}}lcccccccccc@{}}
\toprule
 & \multicolumn{3}{c}{\textbf{All queries}} & \multicolumn{5}{c}{\textbf{Exact match by query type}} & & \\
\cmidrule(lr){2-4} \cmidrule(lr){5-9}
\textbf{Method} & \textbf{Rec.} & \textbf{Prec.} & \textbf{Exact} & \textbf{Obj.} & \textbf{Attr.} & \textbf{Spatial} & \textbf{Set} & \textbf{Reas.} & \textbf{s/query} & \textbf{\textcent/query} \\
\midrule
Memory Over Maps & 70.5 & 67.7 & 51.5 & 78.3 & 80.0 & 36.4 & 29.6 & 27.3 & 1.8 $\pm$ 0.7 & --- \\
ObjMap & 71.1 & 26.1 & 30.1 & 56.5 & 65.0 & 0.0 & 18.5 & 0.0 & $<$0.1 & --- \\
ObjMap + VLM & 75.2 & 74.7 & 46.6 & 69.6 & 70.0 & 40.9 & 33.3 & 0.0 & 2.3 $\pm$ 0.5 & 0.8 \\
TextMap & 73.8 & 81.5 & 68.0 & 82.6 & 65.0 & \textbf{95.5} & 44.4 & 45.5 & 8.5 $\pm$ 7.0 & 1.4 \\
\midrule
\texttt{find\_objects} & 71.8 & 66.0 & 47.6 & 73.9 & 75.0 & 22.7 & 33.3 & 27.3 & 2.4 $\pm$ 0.3 & --- \\
\fido{} & \textbf{97.3} & \textbf{97.3} & \textbf{94.2} & \textbf{91.3} & \textbf{95.0} & \textbf{95.5} & \textbf{92.6} & \textbf{100.0} & 29.5 $\pm$ 18.0 & 3.1 \\
\bottomrule
\end{tabular*}
\end{table*}

We test whether grounding over raw RGB-D scans at query time resolves task-specific object hypotheses better than grounding over representations built before the query. We evaluate over 103 queries ranging from resolving an object category or product; an attribute of it; a spatial relation to another object; a set of more than one target; or reasoning that appearance and geometry alone cannot resolve, such as negation, function, or comparison. We compare against four baselines that ground over a representation built once per scan: Memory Over Maps~\cite{zhou2026memoryovermaps}, ObjMap (after ConceptGraphs~\cite{gu2024conceptgraphs}), ObjMap + VLM, and TextMap; we also ablate \fido{} to \texttt{find\_objects} alone. Predictions are judged manually; Exact is the \% of queries with every target found and no false positives. More details are in Appendix~\ref{app:grounding_vlm}.

\textbf{Result.} \fido{} improves recall, precision, and exact match over every baseline (94.2\% vs.\ 68.0\% exact for TextMap, the best baseline) at higher query time (Table~\ref{tab:grounding}). \fido{} verifies and reasons over the \texttt{find\_objects} tool's outputs and re-invokes it with different labels, using its common-sense reasoning, until it finds the object it needs. We also compare \fido{} against multiple standalone VLMs on the same grounding queries and provide a detailed discussion in the Appendix~\ref{app:grounding_vlm}.

\begin{figure}[t]
\centering
\includegraphics[width=0.95\textwidth]{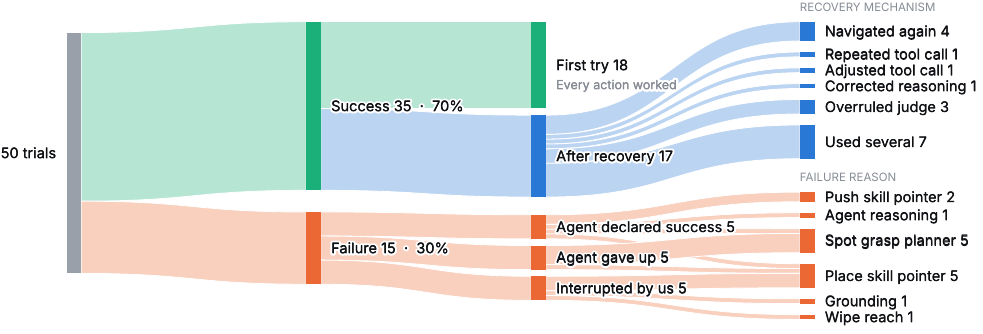}
\caption{%
\textbf{Outcomes of 50 Spot trials.} 17/35 successes recovered from failed tool calls or verifier checks, mostly by re-navigating and retrying. Almost every failure is a robot skill going wrong -- only two are due to grounding or the agent's reasoning.%
}
\label{fig:spot-trials}
\end{figure}

\begin{table}[t]
\centering
\caption{\textbf{Harness ablations.} (a) Missed failures among 20 replayed place and wipe steps that failed in the Spot trials. (b) Successful trials out of five with and without feasibility-aware sampling. (c) Mean $\pm$ std change per instruction over ten paired navigation instructions without \texttt{estimate\_cost}.}
\label{tab:ablations}
\footnotesize
\setlength{\tabcolsep}{3pt}
\renewcommand{\arraystretch}{1.15}
\begin{minipage}[t]{0.31\linewidth}
\vspace{0pt}\centering
\begin{tabular}{@{}lc@{}}
\toprule
\textbf{Evidence shown} & \textbf{Missed} \\
\midrule
Status text only & 9/20 \\
+ before/after images & 6/20 \\
+ verdict (\ours{}) & \textbf{2/20} \\
\bottomrule
\end{tabular}
\subcaption{Effect verification}
\label{tab:ablation_verification}
\end{minipage}%
\hspace{3pt}%
\begin{minipage}[t]{0.33\linewidth}
\vspace{0pt}\centering
\begin{tabular}{@{}lcc@{}}
\toprule
\textbf{Condition} & \makecell{\textbf{Cabinet}\\\scriptsize 1/2 blocked} & \makecell{\textbf{Teddy}\\\scriptsize 2/3 unreach.} \\
\midrule
\ours{} & 5/5 & 5/5 \\
w/o sampler & 2/5 & 1/5 \\
\bottomrule
\end{tabular}
\subcaption{Feasibility-aware sampling}
\label{tab:ablation_feasibility}
\end{minipage}%
\hfill
\begin{minipage}[t]{0.32\linewidth}
\vspace{0pt}\centering
\begin{tabular}{@{}lcc@{}}
\toprule
\textbf{Metric} & \textbf{$\Delta$ per instr.} & \textbf{$p$} \\
\midrule
Distance & $+9.8 \pm 14.7\%$ & $0.039$ \\
Time & $+12.3 \pm 14.3\%$ & $0.020$ \\
\bottomrule
\end{tabular}
\subcaption{Cost-aware planning}
\label{tab:ablation_cost_panel}
\end{minipage}
\end{table}

\subsection{Evaluating the Harness Design}
\label{sec:exp_harness_design}

\subsubsection{Skill Verification}
\label{sec:ablation_skill_verification}

We evaluate whether visual outcome verification (Sec.~\ref{sec:problem_execution}) improves failure detection and recovery. Across the 50 Spot trials, we find 20 place and wipe executions that failed, by hand-checking the verifier's failure verdict. We rebuilt the agent's context up to that turn and the \llm{} for the next tool call, and compare three conditions: (i) status text only, (ii) status text with before and after images, and (iii) the verdict additionally. If it calls observe, one more turn is played with the after image. We see that the agent marked the failed step done and moved on 9 of 20, 6 of 20, and 2 of 20 times as seen in Table~\ref{tab:ablation_verification}. With only the status text, the agent often asked for a picture, and when it did, it usually noticed the failure. When we gave it the pictures anyway, it still missed a third of the failures. The verifier $J$ in \ours{} catches most of the failures, because it asked the question directly and sees nothing else to bias its outcome.

\subsubsection{Feasibility-Aware Sampling}
\label{sec:ablation_sampling}

We evaluate whether removing feasibility-aware sampling of base poses affects skill execution success (Sec.~\ref{sec:problem_sampling}). We evaluate over five trials in two scenarios: 1) putting a measuring tape on a cabinet, where two identical cabinets satisfy the instruction but a chair blocks one of them; and 2) fetching a teddy bear, where three teddy bears are grounded but only one is within reach (the others are blocked by obstacles or high up on a shelf). \ours{}'s harness refuses calls to the navigate tool when it cannot find any satisfying samples. The \llm{} uses this feedback to move to another hypothesis, and every trial ends at a feasible pose as shown in Table~\ref{tab:ablation_feasibility}. Without the sampler, the harness may drive a pose for which the subsequent skill execution fails.

\subsubsection{Cost-Aware Planning and Execution}
\label{sec:ablation_cost}

We evaluate whether cost-aware planning and execution improves execution efficiency. We compare \ours{} with and without \texttt{estimate\_cost} (Sec.~\ref{sec:problem_agent}), keeping the same navigation planner and the same instruction to prefer efficient solutions in both conditions. Without \texttt{estimate\_cost}, the agent drives $9.8 \pm 14.7\%$ further and takes $12.3 \pm 14.3\%$ longer per instruction. It takes a longer route in seven of the ten instructions, slightly shorter in two, and identical in one. We perform a two-sided Wilcoxon signed-rank test over the paired instances, which gives $p=0.039$, and $p=0.02$ (both p-values below $0.05$) over the paired distances and times respectively as shown in Table~\ref{tab:ablation_cost_panel}.

\subsection{Mobile Manipulation on Spot}
\label{sec:experiments_performance}

\begin{table}[t]
\centering
\caption{\textbf{Spot trial results.} Horizon is the number of robot skill calls the task needs (pick, place, push, and wipe, excluding navigate). Progress is the \% of subgoals a failed trial completed. Duration, tool calls, and hosted LLM cost are over every trial, success or failure. All are mean $\pm$ std.}
\label{tab:spot_trials}
\small
\setlength{\tabcolsep}{4pt}
\renewcommand{\arraystretch}{1.15}
\begin{tabular*}{\columnwidth}{@{\extracolsep{\fill}\hspace{\tabcolsep}}lccccc}
\toprule
\textbf{Horizon} & \textbf{Success} & \textbf{Progress (\%)} & \textbf{Duration (min)} & \textbf{Tool calls} & \textbf{Cost (\$)} \\
\midrule
Short (1--2 skills) & 13/16 (81\%) & 50 $\pm$ 43 & 2.5 $\pm$ 1.4 & 12.4 $\pm$ 2.7 & 0.19 $\pm$ 0.13 \\
Medium (3--5 skills) & 12/18 (67\%) & 53 $\pm$ 17 & 4.5 $\pm$ 1.7 & 18.6 $\pm$ 4.7 & 0.48 $\pm$ 0.21 \\
Long (6+ skills) & 10/16 (62\%) & 72 $\pm$ 11 & 8.7 $\pm$ 3.3 & 32.1 $\pm$ 10.0 & 1.09 $\pm$ 0.43 \\
\midrule
All & 35/50 (70\%) & 60 $\pm$ 23 & 5.2 $\pm$ 3.4 & 20.9 $\pm$ 10.3 & 0.58 $\pm$ 0.47 \\
\bottomrule
\end{tabular*}

\vspace{-6pt}
\end{table}

We ran \ours{} on 50 mobile manipulation tasks on Spot, given in plain language: ``clean the whiteboard with the yellow and blue mitt'', ``can you recycle the soda cans? Throw them into a bin'', ``I want all the teddy bears on the yellow toy table, and nothing else. Put the rest away'', and ``someone needs to use the wooden table, can you clear everything off it into a container and bring them the purple cup?''. Together they took 585 skill calls (navigate, pick, place, push, and wipe) and 1,045 agent tool calls. Table~\ref{tab:spot_trials} reports success, task progress, and cost by horizon; Figure~\ref{fig:spot-trials} traces every trial to how it ended, what it recovered from or what failed.

Across the trials, 58\% of wall time was spent inside tool calls (navigate 23\%, place 13\%, pick 9\%, \texttt{find\_objects} 8\%, wipe 3\%, push 2\%) and 41\% inside the agent's LLM calls, about 20 per trial at 7\,s each.
In 25 of the 50 instructions at least one target is not named but has to be worked out, by comparison (``the button with the same color as the smallest chair''), by relation (``the closest container''), or by category (``the unhealthy drinks''); grounding resolved the wrong object in only one trial. In 17 of the 50 trials, the agent recovered after an initial failure, such as wiping the whiteboard again after the initial skill didn't wipe out entire figure, or re-navigating to pick up soda cans which are reflective and the Spot depth sensor didn't pick them up initially.

\section{Conclusion} 
\label{sec:conclusion}

We presented \ours{}, a language-model agent for mobile manipulation that combines task-driven grounding over posed RGB-D observations with an embodiment-aware execution harness. We construct object hypotheses as the instruction requires and use feasibility checks, outcome verification, and execution feedback to guide subsequent actions and plan revisions. Across 103 grounding queries, our approach outperforms the prebuilt semantic representations and the grounding tool without the agent. Our ablations demonstrate the benefits of feasibility-aware sampling and skill execution verification with feedback, while cost-aware planning and execution reduces robot navigation costs. Across 50 real-world Spot trials, \ours{} completed 35 tasks, including 17 successes following recovery from failed tool calls or verification checks. Our failure analysis identifies robot skills as the main remaining bottleneck, while grounding at query time and language-model reasoning add latency. Improving skill reliability and reducing inference overhead remain important directions for extending this approach to broader tasks and environments.

\acknowledgments{We gratefully acknowledge support from NSF grant 2214177; from AFOSR grant FA9550-22-1-0249; from ARO grant W911NF-26-1-A329; from ONR MURI grants N00014-22-1-2740 and N00014-24-1-2603; from the MIT Siegel Family Quest for Intelligence; and from the Robotics and AI Institute.}

\bibliography{references}

\newpage
\appendix
\section*{Appendix}
\section{Grounding Pipeline}
\label{app:grounding_pipeline}

\begin{figure}[!ht]
\centering
\includegraphics[width=\linewidth]{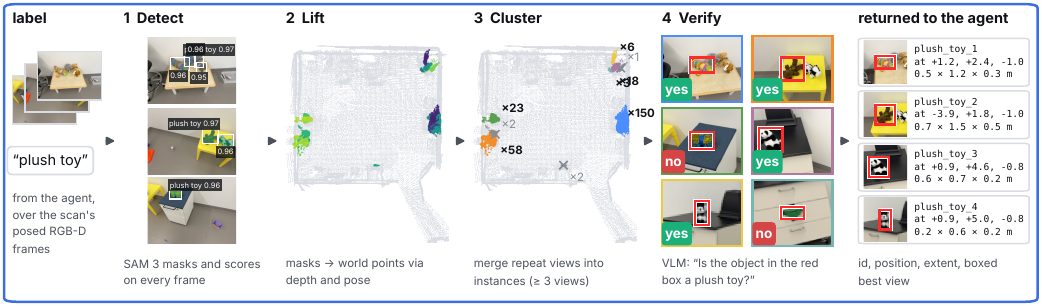}
\caption{\textbf{Task-conditioned grounding.} \fido{} formulates a grounding query and invokes the grounding operator $\mathcal{G}$ over the recorded scan $\mathcal{D}_0$ to generate task-relevant object hypotheses.}
\label{fig:grounding}
\end{figure}

\label{app:grounding_impl}
\fido{} accepts a set of grounding queries ($\mathcal{Q}_t$) and resolves them into a set of object hypotheses $\mathcal{Z}_t$ by searching through the observation memory $\mathcal{D}_0$, using the grounding operator $\mathcal{G}$. We implement $\mathcal{G}$ as the \texttt{find\_objects} tool. 
For each label, we generate bounding boxes, segmentation masks, and detection scores across every recorded RGB image using text-prompted SAM~3~\cite{carion2026sam}.  
To make grounding efficient, we precompute and cache SAM~3's image-encoder features for each recorded RGB image and reuse them across queries. For each segmentation mask, we map it to its corresponding depth image, and back-project its valid depth pixels using the camera intrinsics into the workspace frame using the recorded camera pose. We discard detections with insufficient valid depth. 
For each candidate, we compute its estimated position $\hat{\mathbf p}_j$ as the mean of its accumulated surface points and its spatial extent $\hat{\mathbf d}_j$ as its axis-aligned bounding box. 
For each candidate, we check whether it matches the query label using a locally running Qwen3-VL-8B-Instruct VLM. For each candidate $z_j$, we select a representative view from $\mathcal V_j$, with a bounding box drawn around it, and query whether it matches the label. 
Grounded instances are returned as $\mathcal{Z}_t$, and each instance includes an identifier $\mathrm{id}_j$, its estimated position and extent, viewpoints in which the instance was seen $\mathcal{V}_j$ including the annotated reference image. These outputs allow the agent to inspect the instances against the task and reference it in subsequent tool calls. 

To consolidate repeated observations of the same object, we
process detections in descending score order and greedily merge
their lifted surfaces, independently for each query label.
We represent each surface using 5\,cm voxels and merge it into
an existing cluster when their shared voxels cover at least
5\% of the smaller voxel set. Clusters supported by fewer than
three detections are discarded. For each remaining candidate,
we compute its estimated position $\hat{\mathbf p}_j$ as the
mean of the accumulated surface points and its spatial extent
$\hat{\mathbf d}_j$ as their axis-aligned bounding dimensions.

\section{Open-Ended Grounding: Setup, Baselines, and Standalone VLMs}
\label{app:grounding_vlm}

This appendix gives the full experimental setup, baselines, and metrics for the grounding evaluation in Section~\ref{sec:exp_grounding} (Table~\ref{tab:grounding}), the preprocessing cost of each method, and results for standalone VLMs given every scan frame (Table~\ref{tab:grounding_vlm}).

\textbf{Experimental setup.} We set up four scenes and evaluate our method against four grounding methods on 103 grounding queries. The queries range from resolving an \emph{object} category or product (``sofa'', ``made good granola box''); an \emph{attribute} of it (``black chair'', ``bottle with the blue top''); a \emph{spatial} relation to another object (``water bottle under the sofa'', ``closest chair to the caterpillar toy''); a \emph{set} of more than one target (``plush toys'', ``all the objects on the table with the toy burger''); or \emph{reasoning} that appearance and geometry alone cannot resolve, such as negation, function, comparison, or a reference through another modality (``soda cans not in a container'', ``somewhere to put my organic waste'', ``least ergonomic keyboard'', ``plush toy whose animal is drawn on the whiteboard'').

\textbf{Baselines.} We compare against four ways of grounding a query over a map built before the query is known. Memory Over Maps~\cite{zhou2026memoryovermaps} indexes the scan's keyframes with SigLIP2, shortlists the frames closest to the query, has a VLM (Qwen3-VL-30B) confirm each, and segments and lifts the confirmed frames into 3D instances; we run the authors' code at their settings. ObjMap follows ConceptGraphs~\cite{gu2024conceptgraphs}: an open-vocabulary detector prompted with a fixed vocabulary proposes objects in every frame, their masks are lifted to 3D and merged across views into an object map, and each object carries an embedding of its views; a query returns the objects whose embeddings match it. ObjMap + VLM reads the same map but shortlists the nearest objects and shows each one's stored view to a VLM, keeping those it confirms. TextMap reads the same map serialized as a scene graph, one caption per object with its class and geometry and the spatial relations between objects generated by a VLM, and answers queries via LLM calls over the text. Each of these maps is built once per scan. We run \fido{} and an ablation of \fido{} that only calls the \texttt{find\_objects} tool.

\textbf{Metrics.} Predictions are judged manually against the query's target objects. A duplicate prediction is ignored. Recall and precision are measured over all 148 targets. Exact is the \% of queries with every target found and no false positives.

\textbf{Preprocessing and query cost.} Scan pre-processing on average takes 2\,s for Memory Over Maps to build the SigLIP2 cache, 111\,s to construct the scene graph for ObjMap, 244\,s to construct the scene graph and perform the captioning with parallelized Gemini calls and \$0.85 for TextMap, and 17\,s for \fido{} to encode the SAM3 features. Per query, on average \fido{} costs 3.1 cents, TextMap 1.4, and ObjMap + VLM 0.8.

\textbf{Standalone VLMs.} We now compare \fido{} against standalone VLMs on the same set of grounding queries in Table~\ref{tab:grounding_vlm}. 
For the standalone VLMs, we build the request consisting of the system prompt ("you are given frames of a recorded scan; an object seen in several frames is one object; answer once, in the frame that shows it best"), the raw frames in order, and label them, and the query. We cache the frames once, so the provider uses this cache across the 103 queries on the scans. 
The VLM returns for each object hypothesis, a frame id, a box it draws in that frame, and a few words, and then our harness converts those boxes to pixels, reads the depth, and back-projects it with the frame's pose into one 3D point. 
We argue that grounding on the raw scans rather than grounding from the prebuilt scan is better, and that the standalone VLMs and \fido{} are instances of the same class of methods. 
While \fido{} achieves 94.2\% exact match, outperforming the evaluated prebuilt representations, the best-performing standalone VLM model gemini-3.8-flash achieves 96.1\% at a lower query latency and similar token cost. We have the following takeaways: (i) Sets are the weakest query type for every standalone VLM (865.2 for gemini-3.8-flash vs 92.6 for \fido{}), which is exactly the issue that our 3D clustering handles explicitly. \fido{}'s misses are plain objects and attributes, most likely the detector or verifier dropping object hypothesis; (ii) \fido{} returns what the robot needs next. The VLM's return one point per object without a 3D extent. However, skills consume the extent and views to be feasibility-aware. So, an additional step is needed before robot skills can be executed. \fido{} generates object hypotheses with 3D positions, their extents, and views that they were seen in directly; (iii) The VLMs need the whole scan to fit within its context window. The whole-scan request sends every frame with every request, so context length and token cost scale with the scan, whereas \fido{}'s cost scales with the labels it queries and the instances it inspects. This could become a genuine limitation as workspace size scales. Our scans fit comfortable in context.

\begin{table}[!ht]
\centering
\caption{\textbf{Standalone VLMs on the same grounding queries.} Each model is given every frame of the scan in one request and answers with a box per object, at medium reasoning effort, or off where marked. \fido{} is repeated from Table~\ref{tab:grounding} for reference; it runs on Gemini 3.8 Flash. Query time is mean $\pm$ std, excluding preprocessing; cost is what the hosted model billed, cached input at the cached rate.}
\label{tab:grounding_vlm}
\footnotesize
\setlength{\tabcolsep}{3pt}
\renewcommand{\arraystretch}{1.15}
\begin{tabular*}{\linewidth}{@{\extracolsep{\fill}}lcccccccccc@{}}
\toprule
 & \multicolumn{3}{c}{\textbf{All queries}} & \multicolumn{5}{c}{\textbf{Exact match by query type}} & & \\
\cmidrule(lr){2-4} \cmidrule(lr){5-9}
\textbf{Method} & \textbf{Rec.} & \textbf{Prec.} & \textbf{Exact} & \textbf{Obj.} & \textbf{Attr.} & \textbf{Spatial} & \textbf{Set} & \textbf{Reas.} & \textbf{s/query} & \textbf{\textcent/query} \\
\midrule
Gemini 3.6 Flash (off) & 53.0 & 43.4 & 33.0 & 39.1 & 50.0 & 40.9 & 11.1 & 27.3 & 9.0 $\pm$ 6.7 & 2.6 \\
Gemini 3.6 Flash & 95.3 & 97.9 & 91.3 & \textbf{100.0} & 95.0 & \textbf{100.0} & 77.8 & 81.8 & 21.5 $\pm$ 8.8 & 3.3 \\
Gemini 3.8 Flash & 97.3 & \textbf{100.0} & \textbf{96.1} & \textbf{100.0} & \textbf{100.0} & \textbf{100.0} & 85.2 & \textbf{100.0} & 16.7 $\pm$ 7.0 & 3.2 \\
GPT-5.6 Terra & 75.8 & 81.3 & 65.0 & 82.6 & 75.0 & 86.4 & 33.3 & 45.5 & 12.7 $\pm$ 4.1 & 4.5 \\
GPT-5.6 Sol & 86.6 & 87.8 & 80.6 & 78.3 & 90.0 & 95.5 & 66.7 & 72.7 & 14.1 $\pm$ 5.4 & 9.0 \\
GPT-6 Astra & \textbf{98.0} & 95.4 & 93.2 & \textbf{100.0} & 90.0 & \textbf{100.0} & 85.2 & 90.9 & 17.3 $\pm$ 8.2 & 23.3 \\
\midrule
\fido{} & 97.3 & 97.3 & 94.2 & 91.3 & 95.0 & 95.5 & \textbf{92.6} & \textbf{100.0} & 29.5 $\pm$ 18.0 & 3.1 \\
\bottomrule
\end{tabular*}
\end{table}

\end{document}